\documentclass[runningheads]{llncs}
\usepackage[T1]{fontenc}
\usepackage{graphicx,verbatim}
\usepackage{amsmath}
\usepackage{multirow}
\usepackage{booktabs}
\usepackage{amssymb}
\usepackage{dsfont}
\usepackage{xcolor}
\usepackage{arydshln}
\usepackage{siunitx}
\usepackage{makecell}

\begin{document}
\title{End-to-End Learning of Multi-Organ Implicit Surfaces from 3D Medical Imaging Data}
\titlerunning{End-to-End Implicit Surfaces from 3D Medical Imaging Data}

\renewcommand{\thefootnote}{\fnsymbol{footnote}}
\footnotetext[1]{Corresponding author}
\def\thefootnote{$$}\footnotetext{\textit{This manuscript has been accepted for publication and will be included in the proceedings of ShapeMI 2025.}}

\author{Farahdiba Zarin\inst{1}$^{\star}$ \and
Nicolas Padoy\inst{1,2} \and
Jérémy Dana\inst{2,3,4}
\and
Vinkle Srivastav\inst{1,2}
}
\authorrunning{F. Zarin et al.}
\institute{University of Strasbourg, CNRS, INSERM, ICube, UMR7357, Strasbourg, France \email{fzarin@unistra.fr}\\
\and
Institute of Image-Guided Surgery, IHU Strasbourg, Strasbourg, France
\and
Department of Diagnostic Radiology, McGill University, Montreal, Canada
\and
Augmented Intelligence \& Precision Health Laboratory (AIPHL), McGill University Health Centre Research Institute, Montreal, Canada
}
\maketitle              % typeset the header of the contribution
\begin{abstract}
The fine-grained surface reconstruction of different organs from 3D medical imaging can provide advanced diagnostic support and improved surgical planning. However, the representation of the organs is often limited by the resolution, with a detailed higher resolution requiring more memory and computing footprint. Implicit representations of objects have been proposed to alleviate this problem in general computer vision by providing compact and differentiable functions to represent the 3D object shapes. However, architectural and data-related differences prevent the direct application of these methods to medical images. This work introduces \emph{ImplMORe}, an end-to-end deep learning method using implicit surface representations for multi-organ reconstruction from 3D medical images. \emph{ImplMORe} incorporates local features using a 3D CNN encoder and performs multi-scale interpolation to learn the features in the continuous domain using occupancy functions. We apply our method for single and multiple organ reconstructions using the \emph{totalsegmentator} dataset. By leveraging the continuous nature of occupancy functions, our approach outperforms the discrete explicit representation-based surface reconstruction approaches, providing fine-grained surface details of the organ at a resolution higher than the given input image. The source code will be made publicly available at: \url{https://github.com/CAMMA-public/ImplMORe}

\end{abstract}
\section{Introduction}
% \setlength{\parskip   }{0pt}
% \setlength{\parsep}{0pt}
% % \setlength{\headsep}{30pt}
%  \setlength{\topskip}{0pt}
% \setlength{\topmargin}{0pt}
% \setlength{\topsep}{0pt}
%  \setlength{\partopsep}{0pt}
3D shape reconstruction is a challenging computer vision task, typically tackled by employing explicit shape representations like voxels, polygonal meshes, or point clouds ~\cite{qi2017pointnet,gkioxari2019mesh,wu20153d,wickramasinghe2020voxel2mesh}. These representations however come with inherent limitations such as restricted resolutions (e.g., point clouds, voxels), inability to handle arbitrary topologies (e.g., polygonal meshes, point clouds), and high memory footprints (e.g., voxels), thereby often leading to sub-optimal surface reconstruction quality.

Deep implicit functions have emerged as efficient alternatives for representing 3D shapes and scenes ~\cite{park2019deepsdf,tewari2020state,saito2019pifu,mildenhall2021nerf,chibane2020neural}. The key idea is to represent 3D shape ($\mathcal{S}$) as zero-level set functions using signed distance functions (SDFs) or occupancy functions of a neural network, i.e.,  $\mathcal{S} = \{\mathbf{x} \in \mathbb{R}^3 \ \vert \ f(\mathbf{x};\theta)=0 \}$. Here $f:\mathbb{R}^{3}\times \mathbb{R}^m \rightarrow \mathbb{R} $ is a neural network parameterized by $\theta$ that takes continuous 3D points ($x \in \mathbb{R}^3$) as input, transforms them to higher-dimensional feature space of dimension $m$ and predicts either the signed distance (distance from the 3D surface) or the occupancy value (indicating whether the point is inside or outside the surface). This formulation enables a compact representation of 3D shapes as continuous and differentiable functions. It also offers theoretically infinite resolution to represent 3D shapes, thereby making them suitable for modeling complex shapes with fine-grained geometric details \cite{xie2022neural}. 

In medical imaging, leveraging implicit functions for accurate 3D surface reconstruction of organs or tumors could pave the way to enhance diagnostic support, accurately delineate organs at risk \cite{khan2022implicit}, detect abnormal shapes such as pancreatic fatty infiltration \cite{previtali2023quantitative}, compute surface nodularity for conditions such as liver cirrhosis \cite{elkassem2022multiinstitutional}, and facilitate intra-operative surgical navigation \cite{brunet2019physics}. 

However, employing these techniques in medical imaging encounters two main challenges. Firstly, unlike computer vision datasets, where 3D shapes are readily available in processed high-resolution polygonal meshes \cite{chang2015shapenet}, 3D imaging data exists in volumetric slices, with organs of interest represented as segmented volume slices. This requires the development of end-to-end methods to reconstruct the 3D surfaces of given organs directly from volumetric 3D imaging data. Secondly, these approaches mostly utilize fully connected layers to map continuous 3D points to the signed distance or occupancy values from a single class of objects, hindering their scalability to different classes using the same neural network. Moreover, using the globally fully connected layers neglects incorporating local information from the input data, potentially leading to overly smooth surface reconstructions. The recent applications of implicit functions in medical imaging have mainly focused on improving the segmentation performance \cite{marimont2022implicit,khan2022implicit}, refinement of segmentation masks \cite{yang2022neural}, and learning topology-preserving implicit shape templates \cite{sun2022topology} using non-end-to-end design. Explicit representation-based methods such as Voxel2mesh \cite{wickramasinghe2020voxel2mesh} and its organ-specific variants \cite{kong2021deep,bongratz2022vox2cortex} provide end-to-end design but exploit explicit organ-specific priors for the surface reconstruction. 

This work introduces \emph{ImplMORe}, Implicit Multi-Organ Reconstruction, a deep-learning approach for end-to-end multi-organ implicit surface reconstruction from 3D medical imaging. We take inspiration from implicit feature network (IFNet) \cite{chibane2020implicit} and convolutional occupancy functions \cite{peng2020convolutional} to incorporate local features for implicit shape learning. We utilize a 3D CNN encoder to obtain interpolated multi-scale features at the sampled 3D query points. The interpolated feature points are then passed through a decoder network to learn the occupancy function. The interpolation of the features at multiple scales helps to promote learning in the continuous domain. To enable the learning of multiple organs in function space, we introduce a multi-head implicit decoder with a decoder for each organ. This design enables the mitigation of the issue of the same coordinate between organs in contact sharing their occupancy values, especially when images are downsampled. We further propose a patch-wise learning approach to process the high-dimensional images using the same 3D CNN encoder to extract the multi-scale features. Processing the high-dimensional images in 3D patches provides the benefits of finer reconstruction, especially for smaller organs. We also enhance the inductive biases of the 3D CNN encoder by applying the same affine augmentation to the input images and the sampled query points. Finally, we propose to sample the query points to use occupancy values generated from the high-resolution meshes during pre-processing rather than the corresponding ground truth segmentation masks where resolution restricts the quality of the generated occupancy. Evaluation of \emph{ImplMORe} on the \emph{totalsegmentator} dataset \cite{wasserthal2022totalsegmentator}, along with extensive ablation studies, show the effectiveness of our approach against discrete and explicit representation-based surface reconstruction approaches. We summarize our contributions as follows:
\begin{itemize}
\setlength\itemsep{0.3mm}
\item Propose end-to-end 3D organ reconstruction using implicit surface representation directly from 3D medical image data.
\item Reconstruction from low-resolution whole images and patch-wise reconstruction from high-resolution images for variable image resolutions.
\item Image-point paired data augmentation to strengthen inductive bias of the 3D feature encoder and for retaining spatial relationship between augmented images and sampled query points.
\end{itemize}
% 1) sampling 2) data augmentation   
% %To eliminate any level of dependency on the location of the query points themselves and emphasize learning from features only, IF-Nets utilized only the features of the query points themselves to obtain their occupancy and excessively sample around the query points to obtain the features shared by those neighboring points 
% In this paper, we achieve end-to-end multi-organ 3D implicit reconstruction that is non-reliant on prior anatomical information with minimal architectural changes across diverse datasets with varying image dimensions and anatomical targets represented by occupancy functions. 
\section{Methodology}
\subsection{Query Point Sampling from Data Processing}
We sample the continuous 3D query points from two sources: ground-truth meshes and input image volume. The ground truth meshes are obtained by applying marching-cube \cite{lorensen1998marching} on the segmented high-resolution volume slices. The points are then sampled from the mesh vertices. The sampled points are displaced by a small factor randomly sampled from a standard distribution of mean 0 and deviation of 0.1 to include distant voxels and 0.01 to include points on the interior and exterior of the target organ in proximity to the organ surface. Points from the input image volume are obtained by randomly selecting voxels and discarding the image background for reduced negative samples. After selecting the 3D query points from the two sources, we follow implicit waterproofing \cite{chibane2020implicit} to compute the Occupancy values [0, 1], where points outside the boundary of the target organ are assigned 0, and those inside are assigned 1.

\subsection{Model Architecture}
\begin{figure*}[h!]
\centering
\begin{center}
    \includegraphics[width=1.\linewidth]{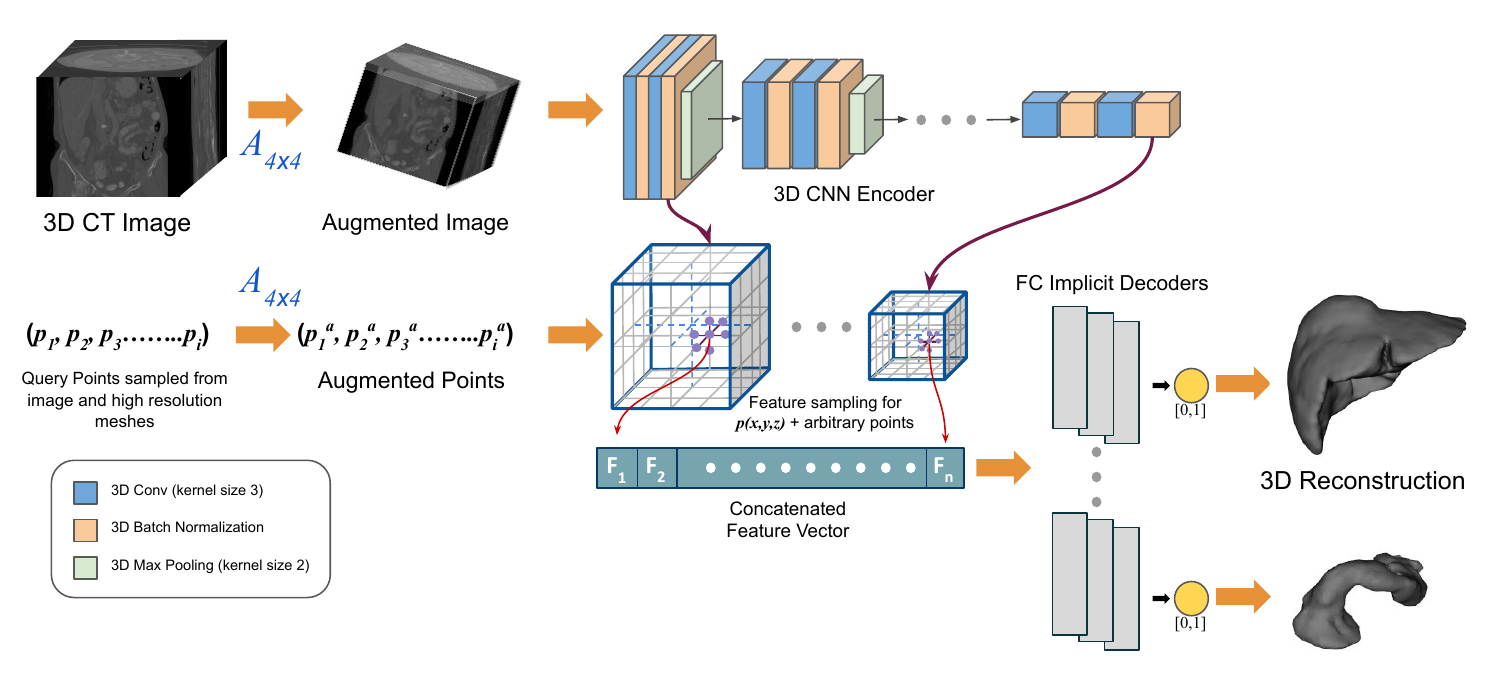}
    \caption{End-to-end implicit multiorgan reconstruction pipeline.}
    \label{multiorganpipeline}
\end{center}
\end{figure*}

The overall architecture of \emph{ImplMORe} is illustrated in Figure \ref{multiorganpipeline} comprising a 3D CNN encoder and fully connected implicit decoders. The 3D CNN encoder receives the whole 3D volume as input. The fully connected implicit decoders then utilize the features from the 3D CNN encoder to differentiate between points inside and outside the organ for a set of input 3D query coordinates. The network outputs the occupancy value based solely on the features provided by the encoder for the specific query point.
% that receives the whole 3D volume as input
% In order to compute occupancy values with no prior awareness of shape or organ location, the network as illustrated in Figure \ref{multiorganpipeline} comprises a CNN encoder that receives the whole 3D volume as input. The fully connected implicit decoder that then follows utilizes the features from this encoder to differentiate between points inside and outside the organ for a set of 3D query coordinates. The network outputs the occupancy value based solely on the features provided by the encoder for the specific query point. Encoder learning rate is of 1x10\textsuperscript{-4}, and 1x10\textsuperscript{-8} for decoder with Adam optimizer.

\subsubsection{CNN Encoder}

Given the proven capabilities of the features obtained from convolution neural networks in segmentation and classification tasks applied to 2D and 3D medical images, the encoder architecture for feature extraction is inspired by the UNet encoder \cite{ronneberger2015u}. This encoder comprises five CNN blocks, each employing 3D CNN layers and subsequent downsampling of images at each layer. This design ensures the extraction of local and global features from initial to final layers. Within each CNN block, there are two sets of 3x3 convolution kernels followed by batch normalization. After each max pool layer, the feature maps undergo a channel doubling while halving in spatial dimensions.

% mention layer as well as feature maps

\subsubsection{FC Decoder}

The implicit decoder is a series of 3 fully connected 1D convolution layers that take all the queried features for each input coordinate and output an occupancy value based on the features. The decoder takes not only the query points in the training data as input, but also generates surrounding points in the vicinity, enabled by its continuous feature sampling capabilities irrespective of the feature map size. Features obtained from the output of each 3D CNN encoder block are sampled from a continuous grid, allowing the decoder to learn by disseminating both local and global features. 

Query 3D coordinates are used solely for feature sampling, and only sampled features $F_1(I_{\mathbf{p}}) \times F_2(I_{\mathbf{p}}) \times \ldots \times F_n(I_{\mathbf{p}})$ for each point $\mathbf{p} \in \mathbb{R}^3$ in the image $I$ are aggregated and provided to the FC layers, which is a function of the features extraction from the image $I$ at query point locations $\mathbf{p}$. This leverages the implicit learning capabilities of the decoder without the influence of definitive explicit features.

The decoder for the multi-organ architectures is of two variants. In the first variant, a common decoder is employed to distinguish organs based on their features, while the second variant, called the multi-decoder approach, uses a dedicated decoder for each organ. The output layer for the single-organ decoder and the multiple decoders in the multi-organ provides the probability of the points belonging inside or outside the organ, while the multi-organ single decoder pipeline outputs $\mathbf{0..c}$ for $\mathbf{c}$ number of organs where 0 is for points belonging to the background.

\subsubsection{Loss Function}

In the single-organ variant of the architecture and the multi-organ variant with a dedicated decoder for each organ, generated occupancy is in the range [0,1]. The loss to be minimized for training the model is a binary cross entropy (BCE) computed between the ground truth and the predicted occupancy, defined as follows:

% %\[
% \begin{multline}
% \text{{$L_{BCE}$}}(\textbf{y}, y_{\hat{}}) = -\frac{1}{b \cdot n} \sum_{i=1}^{b} \sum_{j=1}^{n} \left( \textbf{y}(i,j) \cdot \log\left(\sigma(y_{\hat{}}(i, j))\right) \right.
% \left. + (1 - \textbf{y}(i, j)) \cdot \log\left(1 - \sigma(y_{\hat{}}(i, j))\right) \right)
% %\]
% \end{multline}\label{eqBCEloss1}
%\[

% \begin{equation}
\begin{multline}
\text{{$L_{BCE}$}}(\textbf{y}_c, y_{\hat{}_c}) = \sum_{k=1}^{c} ( -\frac{1}{b \cdot n} \sum_{i=1}^{b} \sum_{j=1}^{n} \left( \textbf{y}_k(i,j) \cdot \log\left(\sigma(y_{\hat{}_k(i, j)})\right)\right) \\
+ (1 - \textbf{y}_k(i, j)) \cdot \log(1 - \sigma(y_{\hat{}_k(i, j)})) )\
% \]
\end{multline}
% \end{equation}
\label{eqBCEloss2}

The sigma (\(\sigma\)) function is applied element-wise to the predicted probabilities of the occupancy \(y_{\hat{}}(i, j)\) being 1. The outer summation iterates over the batch size (\(b\)), while the inner summation iterates over the number of sampled query points (\(n\)). The binary cross-entropy (BCE) loss is computed for each element in the batch and query points and then averaged over the total number of elements (\(b \cdot n\)). For multiple decoders, the BCE loss of each decoder output for \textbf{c} number of organs is summed. The single decoder variant for multiple organs is trained using a multi-class cross-entropy loss.

\section{Experimental Setup}

\subsection{Data Processing}
To confirm the ability of the model to successfully reconstruct organs from various medical images of different sizes with varying anatomical focus areas, the \emph{totalsegmentator} dataset \cite{wasserthal2022totalsegmentator} is used for the evaluation. The liver organ is selected for the single-organ pipeline, and the liver and the pancreas are selected for the multi-organ variant to evaluate both large and small organs. Given the absence of the desired organs in many of the images of this diverse dataset, images containing the entire liver or most of the liver are retained in the final dataset, resulting in a final dataset of 581 CT volumes,
split into 291 for training, 116 for validation, and 174 for testing.

3D images are normalized with the mean and standard deviation of the pixel values and padded to match the largest image dimension of 500$^{3}$. Meshes are generated from the ground truth segmentation, followed by smoothing and vertex normalization to remove staircase effects between slices while retaining surface curvature details. For training using whole images at a lower resolution, images are downsampled to 128$^{3}$. For the single organ model, 300k query points are sampled - 200k from the image and 100k from the mesh boundaries. For the multi-organ model, 400k query points are sampled, 200k from the images, and 200k from the mesh boundaries (100k query points for each organ).
%Single organ has 200k query points sampled from images. 100k query points sampled from mesh boundary of each organs were added for the multiorgan, bringing the total to 400k.
%Query points are generated from the normalized sampled coordinates of these processed images, with boundary points and occupancy values being generated from the high resolution meshes.
%for mesh processing on a constant cubical 3D grid
To improve the robustness of the model on unseen data and to strengthen the inductive bias of the 3D CNN encoder, affine geometric augmentation ($A_{4x4}$) consisting of random translation, rotation, and scaling is applied on the input data and the corresponding 3D query points, as illustrated in as in Figure \ref{multiorganpipeline}.

% augmentation employing random translation, rotation, and scaling of the images is performed during training. Generating augmented data involves the affine transformation of not just the input images but query points to maintain spatial correspondence .

%registration of the query points from the moving to fixed image to maintain spatial correspondence as in Fig \ref{multiorganpipeline}. We elaborate this further for high resolution images.

Training and inference for the low-resolution images use the whole image as inputs and generate occupancy for a uniform grid sampling the entire image at a higher resolution of 256$^{3}$. Explicit forms of the occupancy are generated using marching cubes \cite{lorensen1998marching} for further visualization and evaluation.

\subsubsection{High Resolution Image}
To leverage the same architecture with minimal changes, patch-wise training is employed for the high-resolution images of 500$^{3}$. 6000k points are sampled from the high-resolution images. The same number of additional boundary points sampled from meshes of each organ are added for the multi-organ variant model. Random patches of dimensions 128$^{3}$ are generated from the input image during training. Patches are not used to discard excessive background patches in the image if the highest intensity voxel is the same as the minimum intensity in the given patch. 

%An alternate method references the voxel ground truth segmentation masks, discarding patches if the corresponding segmentation masks of the patches are empty. (not mentioned in results so skipped, though experiment exists)

Query points and ground truth occupancy values for points within the bounding volume generated by the patches are extracted and translated to comply with a new coordinate system referencing the patch as an independent image with its own grid.

Patch-wise inference generates overlapping patches of 128$^{3}$ from the whole image of 500$^{3}$ following boundary padding to 512$^{3}$ to maintain the same dimension for all generated patches. The patch-wise inference is performed, and probabilities of overlapping patches are smoothed out using a 3D hann window function \cite{oppenheim1999discrete}. Generated occupancy values are obtained over a grid of 1024$^{3}$.

\section{Results and Discussion}

\subsection{Single-Organ and Multi-Organ Implicit Reconstruction}

\begin{table}[]
\caption{Proposed single organ and multi-organ implicit reconstruction pipelines. %Resolution is 128\textsuperscript{3} for all except Ours\textsubscript{256}[Patch]. 
Hidden dimensions are 512 for single organ and 256 for multi-organ. Data augmentation is present for all methods except \emph{ImplMORe} and \emph{ImplMORe(Seg)}.}

\centering
\setlength{\tabcolsep}{2.7pt}
\scalebox{0.7}{\begin{tabular}{cc|cccccccc}
\toprule
\cline{1-10}
& \textbf{Method}        & \multicolumn{2}{c}{\textbf{HD}}          & \multicolumn{2}{c}{\textbf{ASSD}}       & \multicolumn{2}{c}{\textbf{IOU}}        & \multicolumn{2}{c}{\textbf{CD}}         \\ 
\cline{3-10}
 &  & \multicolumn{8}{c}{Liver}                                                                 
 \\\midrule

\parbox[c]{5mm}{\multirow{5}{*}{\rotatebox[origin=c]{90}{\makecell{single organ}}}}

 & UNet                   & \multicolumn{2}{c}{10.4±4.8}            & \multicolumn{2}{c}{4.9±1.4}           & \multicolumn{2}{c}{76.7±4.2}           & \multicolumn{2}{c}{0.1±0.3}           \\
 & Voxel2Mesh             & \multicolumn{2}{c}{12.3±10.8}            & \multicolumn{2}{c}{5.4±2.0}           & \multicolumn{2}{c}{83.3±5.4}           & \multicolumn{2}{c}{0.7±1.5}           \\
 & Voxel2Mesh*            & \multicolumn{2}{c}{9.4±11.7}            & \multicolumn{2}{c}{4.4±4.4}           & \multicolumn{2}{c}{86.6±7.5}           & \multicolumn{2}{c}{0.8±5.7}           \\
 & \emph{ImplMORe(Seg)}           & \multicolumn{2}{c}{8.7±3.0}            & \multicolumn{2}{c}{4.3±1.1}           & \multicolumn{2}{c}{79.3±3.3}           & \multicolumn{2}{c}{0.1±0.2}           \\
% \multicolumn{1}{l|}{} 
& \emph{\textbf{ImplMORe}}      & \multicolumn{2}{c}{\textbf{6.5±3.6}}   & \multicolumn{2}{c}{\textbf{2.6±1.0}}  & \multicolumn{2}{c}{\textbf{86.9±4.7}}  & \multicolumn{2}{c}{\textbf{0.03±0.04}}  \\\midrule
\cline{1-10}
&  & \multicolumn{2}{c}{\textbf{HD}}          & \multicolumn{2}{c}{\textbf{ASSD}}       & \multicolumn{2}{c}{\textbf{IOU}}        & \multicolumn{2}{c}{\textbf{CD}}         \\  
\cline{3-10}
  &                     & Liver               & Pancreas           & Liver              & Pancreas           & Liver              & Pancreas           & Liver              & Pancreas           \\\midrule
\parbox[c]{5mm}{\multirow{3}{*}{\rotatebox[origin=c]{90}{\makecell{multi}}}}
\parbox[c]{5mm}{\multirow{3}{*}{\rotatebox[origin=c]{90}{\makecell{organ}}}}
& UNet                   & 20.4±41.2           & 67.8±28.7          & 7.5±10.5          & 18.6±11.9          & 72.3±7.0          & 8.1±11.3          & 0.8±6.0          & 1.8±2.2          \\
& \emph{ImplMORe}               & 6.6±2.5           & 16.0±10.9          & 2.7±0.9          & 5.6±2.6          & 86.7±4.1          & 34.2±15.1          & 0.03±0.1          & 0.14±0.2          \\
& \emph{\textbf{ImplMORe-Patch}} & \textbf{7.0±10.5} & \textbf{9.3±9.2} & \textbf{3.1±1.3} & \textbf{3.3±1.7} & \textbf{84.2±3.8} & \textbf{53.8±14.0} & \textbf{0.06±0.2} & \textbf{0.08±0.2} \\ \bottomrule
\end{tabular}}

\label{tab:both-results}
\end{table}

\begin{figure*}[hbt!]
\centering
\begin{center}
    \includegraphics[width=1\linewidth]{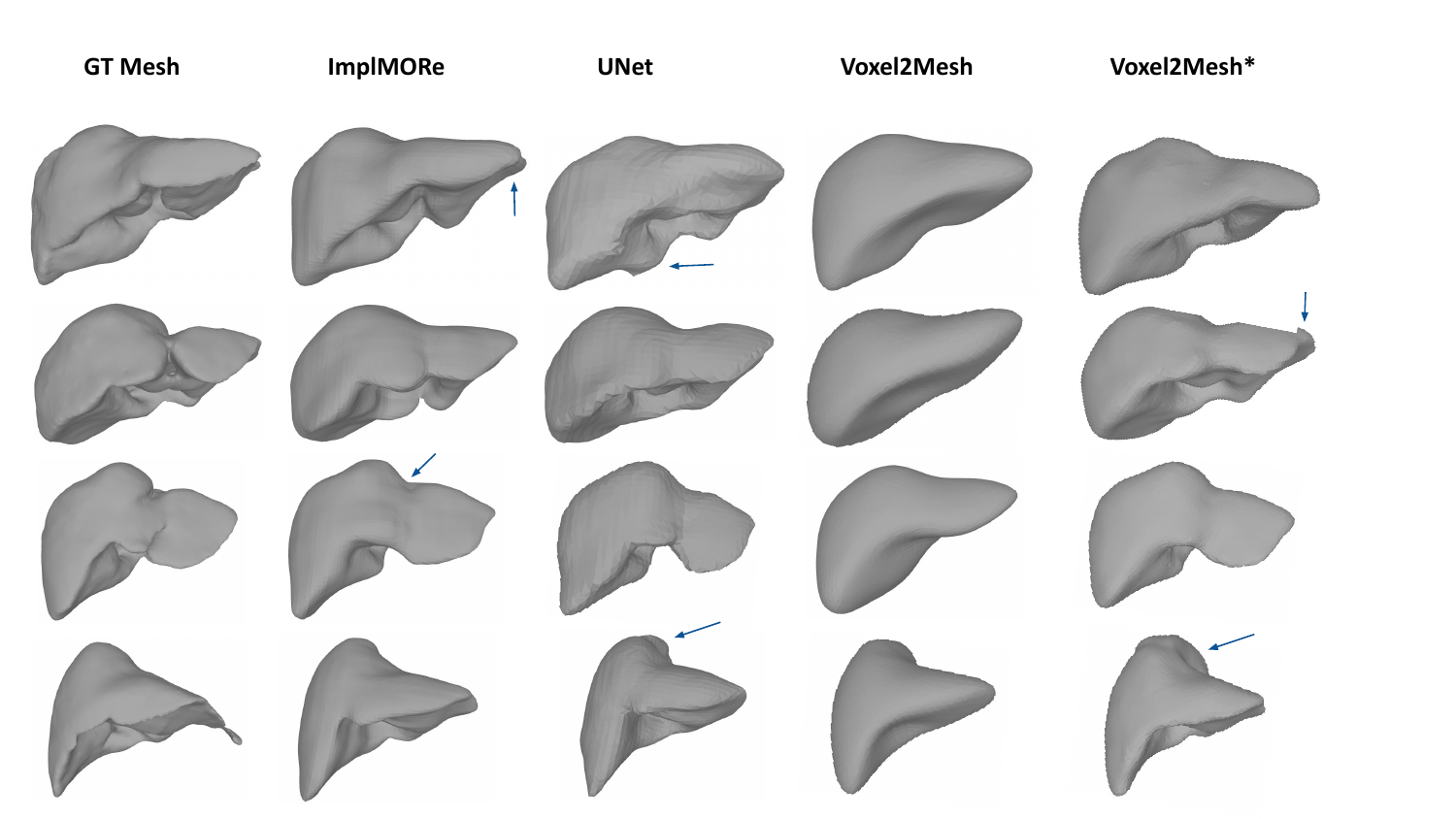}
    \caption{Qualitative results of single-organ pipelines from Table \ref{tab:both-results} compared with explicit baselines. Arrows for \emph{ImplMORe} indicate the successful capturing of details by the implicit network. For Voxel2Mesh \cite{wickramasinghe2020voxel2mesh}, the original architecture smoothed all details and, hence, artifacts. The artifacts are more pronounced when the smoothing is reduced, as indicated by arrows for the modified Voxel2Mesh* and for UNet. Our pipeline captured greater detail and reduced artifacts produced by the explicit pipelines. Despite all presented models being trained on lower-resolution CT volumes, our implicit network can work with low-quality images for large organs.}
    \label{appendix_img_single}
\end{center}
\end{figure*}

Table \ref{tab:both-results} shows the results for the single-organ and multi-organ pipelines. We compare the results using the following metrics: 90\% Hausdorff Distance (HD), Average Symmetric Surface Distance (ASSD) in mm, Chamfer Distance (CD) using factor x10\textsuperscript{-3}, and Intersection over Union (IOU) in \%. The baseline method consists of 3D UNet segmentation architecture \cite{ronneberger2015u} and Voxel2Mesh \cite{wickramasinghe2020voxel2mesh}. We convert the output segmentation map of the 3D UNet into the 3D surfaces using marching cubes \cite{lorensen1998marching} for the evaluation. Results of two versions of Voxel2Mesh \cite{wickramasinghe2020voxel2mesh} for mesh generation directly from 3D medical images are also reported for the single organ. The modified version Voxel2Mesh* in Table \ref{tab:both-results} has lowered regularization loss (normal loss reduced 100-fold) to promote detailed surface generation to ensure comparability. \emph{ImplMORe(Seg)} denotes the proposed pipeline with occupancy values generated from the ground truth segmentation masks. The rest of our pipelines use occupancy from high-resolution ground truth meshes and input images. As illustrated in Table \ref{tab:both-results}, generation of occupancy from high-resolution meshes in \emph{ImplMORe} proved vital in outperforming the UNet and the other baselines. This could be attributed to the lack of subvoxel accuracy when using the segmentation ground truths as the source of occupancy due to the resolution being a major limitation.

\begin{figure*}[hbt!]
\centering
\begin{center}
    \includegraphics[width=.9\linewidth]{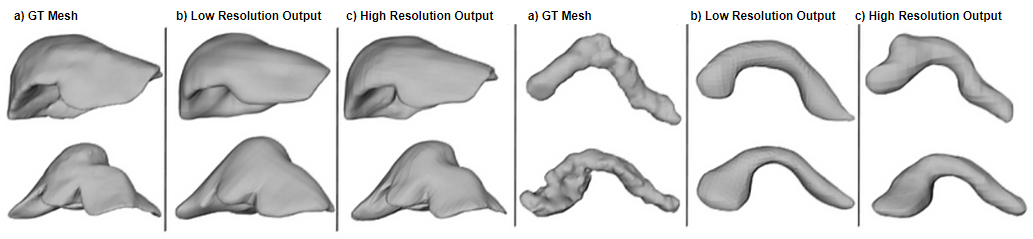}
    \caption{Outputs of the proposed \emph{ImplMORe} and \emph{ImplMORe-Patch} pipelines, where a) is the ground truth mesh, b) is the result using a low-resolution whole image, and c) the result of using patches from the high-resolution image.}
    
    \label{gt_low_high_res}
\end{center}
\end{figure*}

Multi-organ pipelines \emph{ImplMORe} and \emph{ImplMORe-Patch} maintain the liver performance while implicitly showcasing our model's ability to represent multiple organs. The \emph{ImplMORe-Patch} however did not result in noticeable improvements for the liver. The drastic decrease in HD, ASSD, CD, and increase in IOU for the pancreas in Table \ref{tab:both-results} highlights the major benefit of the application of patches for smaller organs. Larger organs, such as the liver in Figure \ref{gt_low_high_res}, exhibit finer surface details in  \emph{ImplMORe-Patch}. Small organs such as the pancreas exhibit notable improvements from the patch-wise approach, where the shape is represented more accurately.

\subsection{Implicit Decoder Configuration for Multi-organ pipeline}
\setlength{\tabcolsep}{.51pt}
\begin{table*}[hbt!]
\caption{Multi-organ pipeline with liver and pancreas trained using low-resolution whole images (128\textsuperscript{3}) with ablation on decoder configurations. Models are trained with data augmentation and L2 regularization.}
\centering
\scalebox{0.8}{
\begin{tabular}{c|c|cccccccc}
\toprule
\multirow{2}{*}{\textbf{\begin{tabular}[c]{@{}c@{}}Hid\\ Dims\end{tabular}}} & \multirow{2}{*}{\textbf{\begin{tabular}[c]{@{}c@{}}Dec\\ Type\end{tabular}}} & \multicolumn{2}{c}{\textbf{HD}} & \multicolumn{2}{c}{\textbf{ASSD}} & \multicolumn{2}{c}{\textbf{IOU}} & \multicolumn{2}{c}{\textbf{CD}} \\ \cline{3-10} 
 &  & Liver & Pancreas & Liver & Pancreas & Liver & Pancreas & Liver & Pancreas \\ \midrule
512 & single & 60.2±92.5 & 32.0±39.0 & 13.3±15.0 & 11.0±9.4 & 70.2±11.1 & 25.0±01.0 & 03.1±06.7 & 01.0±03.0 \\ 
256 & single & 88.3±101.9 & 138.1±200.2 & 21.9±20.5 & 31.0±32.9 & 57.8±12.2 & 18.3±07.2 & 05.4±10.9 & 18.2±31.4 \\ 
512 & multi & 06.8±02.3 & 16.7±10.9 & \textbf{02.7±01.1} & 05.8±02.6 & \textbf{86.7±03.3} & 32.8±14.4 & 0.05±0.27 & \textbf{0.14±0.19} \\ 
256 & multi & \textbf{06.6±02.5} & \textbf{16.0±10.9} & \textbf{02.7±00.9} & \textbf{05.6±02.6} & \textbf{86.7±04.1} & \textbf{34.2±15.1} & \textbf{0.03±0.06} & \textbf{0.14±0.19} \\ \bottomrule

\end{tabular}}

\label{tab:multiorgan-results}
\end{table*}

Table \ref{tab:multiorgan-results} shows the results for the proposed multi-organ reconstruction pipeline using two variants of implicit decoder configurations. For the first variant, utilizing a single decoder architecture and cross-entropy loss, we observe a greater tendency of the model to overfit on the training data and produce artifacts in the output shapes. We observe much more stable training and improved results for the second variant, utilizing a separate decoder for each organ and binary cross-entropy loss. Considering the performance of the individual organs, no degradation of the performance was observed for the liver compared to its results from the single-organ pipeline.

\subsection{Effect of Image-Point Data Augmentation}
\setlength{\tabcolsep}{.7pt}
\begin{table*}[h!]
\caption{Influence of data augmentation on multi-organ pipeline for liver and pancreas for low resolution whole images.}
\centering
\scalebox{0.8}{
\begin{tabular}{c|c|cccccccc}
\toprule
\multirow{2}{*}{\textbf{\begin{tabular}[c]{@{}c@{}}Hid \\ Dims\end{tabular}}} & \multirow{2}{*}{\textbf{\begin{tabular}[c]{@{}c@{}}Data \\ Aug\end{tabular}}} & \multicolumn{2}{c}{\textbf{HD}} & \multicolumn{2}{c}{\textbf{ASSD}} & \multicolumn{2}{c}{\textbf{IOU}} & \multicolumn{2}{c}{\textbf{CD}} \\ \cline{3-10} 
 &  & Liver & Pancreas & Liver & Pancreas & Liver & Pancreas & Liver & Pancreas \\ \midrule
512 &  & 9.23±7.19 & 21.0±15.5 & 3.40±1.89 & 7.02±4.57 & 83.5±06.8 & 27.7±14.7 & \textbf{0.06±0.13} & 0.24±0.43 \\ 
512 & \checkmark & 8.83±17.2 & 16.4±11.2 & 2.98±1.85 & 5.92±2.96 & 86.4±03.4 & \textbf{34.0±14.3} & 0.10±0.46 & 0.15±0.22 \\ 
256 &  & 8.75±7.83 & 40.2±93.5 & 3.22±2.11 & 23.3±82.2 & 84.4±07.5 & 25.2±16.0 & 0.06±0.17 & 15.8±80.4 \\ 
256 & \checkmark & \textbf{7.76±16.8} & \textbf{15.9±10.7} & \textbf{2.73±2.00} & \textbf{5.70±2.65} & \textbf{87.0±3.29} & 32.3±15.0 & 0.09±0.78 & \textbf{0.14±0.22} \\ \bottomrule
\end{tabular}}

\label{tab:augmentation-ablation}
\end{table*}

Our data augmentation transforming the query points to the augmented image coordinate system for maintaining spatial relation is evaluated on the multi-decoder variants of different hidden dimensions in Table \ref{tab:augmentation-ablation}. Both configurations exhibited significantly improved HD and ASSD for both organs.

\section{Conclusion}

In this paper, we present  \emph{ImplMORe}, an end-to-end deep-learning pipeline for 3D implicit reconstruction of multiple organs directly from 3D medical images without prior anatomical knowledge. Our approach consists of a 3D CNN encoder to extract interpolated multi-scale features and multiple implicit decoders to learn the organ shapes in function space. We improve the performance using image-point pair data augmentation by applying the same affine augmentation on the input 3D volume and the sampled query points, ensuring spatial correspondence in the image and the query points. We further learn high-quality features, particularly for small organs, by training the model on high-resolution patch volumes. 
%To the best of our knowledge, our work is the first to introduce an end-to-end implicit surface reconstruction pipeline for image-to-organ reconstruction for 3D medical imaging. %with improved performance by utilizing image-point pair data augmentation for medical images.

% further enhances our training with a patch-wise pipeline designed to utilize occupancy values generated from high-resolution mesh while maintaining spatial correspondence. We achieve improvements with minimal architectural changes for different image resolutions as a benefit of the patch-wise approach, which maintains the performance for large organs and drastically improves for smaller organs. 

% Acknowledgments---Will not appear in anonymized version
%\midlacknowledgments{We thank a bunch of people.}

\begin{credits}
\subsubsection{\ackname} This research was supported by French state funds managed within the ‘Plan Investissements d’Avenir’ by the ANR under references ANR-21-RHUS-0001 (RHU DELIVER) and ANR-10-IAHU-02 (IHU Strasbourg). This work was granted access to the servers/HPC resources managed  by CAMMA, IHU Strasbourg, Unistra Mesocentre, and GENCI-IDRIS [Grant 2021-AD011011638R3, 2021-AD011011638R4].

\subsubsection{\discintname}
The authors have no competing interests to declare that are relevant to the content of this article.
\end{credits}
\bibliographystyle{splncs04}
\bibliography{mybibliography}

\begin{thebibliography}{10}
\providecommand{\url}[1]{\texttt{#1}}
\providecommand{\urlprefix}{URL }
\providecommand{\doi}[1]{https://doi.org/#1}

\bibitem{bongratz2022vox2cortex}
Bongratz, F., Rickmann, A.M., P{\"o}lsterl, S., Wachinger, C.: Vox2cortex: fast explicit reconstruction of cortical surfaces from 3d mri scans with geometric deep neural networks. In: Proceedings of the IEEE/CVF Conference on Computer Vision and Pattern Recognition. pp. 20773--20783 (2022)

\bibitem{brunet2019physics}
Brunet, J.N., Mendizabal, A., Petit, A., Golse, N., Vibert, E., Cotin, S.: Physics-based deep neural network for augmented reality during liver surgery. In: Medical Image Computing and Computer Assisted Intervention--MICCAI 2019: 22nd International Conference, Shenzhen, China, October 13--17, 2019, Proceedings, Part V 22. pp. 137--145. Springer (2019)

\bibitem{chang2015shapenet}
Chang, A.X., Funkhouser, T., Guibas, L., Hanrahan, P., Huang, Q., Li, Z., Savarese, S., Savva, M., Song, S., Su, H., et~al.: Shapenet: An information-rich 3d model repository. arXiv preprint arXiv:1512.03012  (2015)

\bibitem{chibane2020implicit}
Chibane, J., Alldieck, T., Pons-Moll, G.: Implicit functions in feature space for 3d shape reconstruction and completion. In: Proceedings of the IEEE/CVF conference on computer vision and pattern recognition. pp. 6970--6981 (2020)

\bibitem{chibane2020neural}
Chibane, J., Pons-Moll, G., et~al.: Neural unsigned distance fields for implicit function learning. Advances in Neural Information Processing Systems  \textbf{33},  21638--21652 (2020)

\bibitem{elkassem2022multiinstitutional}
Elkassem, A.A., Allen, B.C., Lirette, S.T., Cox, K.L., Remer, E.M., Pickhardt, P.J., Lubner, M.G., Sirlin, C.B., Dondlinger, T., Schmainda, M., et~al.: Multiinstitutional evaluation of the liver surface nodularity score on ct for staging liver fibrosis and predicting liver-related events in patients with hepatitis c. American Journal of Roentgenology  \textbf{218}(5),  833--845 (2022)

\bibitem{gkioxari2019mesh}
Gkioxari, G., Malik, J., Johnson, J.: Mesh r-cnn. In: Proceedings of the IEEE/CVF international conference on computer vision. pp. 9785--9795 (2019)

\bibitem{khan2022implicit}
Khan, M.O., Fang, Y.: Implicit neural representations for medical imaging segmentation. In: Medical Image Computing and Computer Assisted Intervention--MICCAI 2022: 25th International Conference, Singapore, September 18--22, 2022, Proceedings, Part V. pp. 433--443. Springer (2022)

\bibitem{kong2021deep}
Kong, F., Wilson, N., Shadden, S.: A deep-learning approach for direct whole-heart mesh reconstruction. Medical image analysis  \textbf{74},  102222 (2021)

\bibitem{lorensen1998marching}
Lorensen, W.E., Cline, H.E.: Marching cubes: A high resolution 3d surface construction algorithm. In: Seminal graphics: pioneering efforts that shaped the field, pp. 347--353 (1998)

\bibitem{marimont2022implicit}
Marimont, S.N., Tarroni, G.: Implicit u-net for volumetric medical image segmentation. In: Medical Image Understanding and Analysis: 26th Annual Conference, MIUA 2022, Cambridge, UK, July 27--29, 2022, Proceedings. pp. 387--397. Springer (2022)

\bibitem{mildenhall2021nerf}
Mildenhall, B., Srinivasan, P.P., Tancik, M., Barron, J.T., Ramamoorthi, R., Ng, R.: Nerf: Representing scenes as neural radiance fields for view synthesis. Communications of the ACM  \textbf{65}(1),  99--106 (2021)

\bibitem{oppenheim1999discrete}
Oppenheim, A.V.: Discrete-time signal processing. Pearson Education India (1999)

\bibitem{park2019deepsdf}
Park, J.J., Florence, P., Straub, J., Newcombe, R., Lovegrove, S.: Deepsdf: Learning continuous signed distance functions for shape representation. In: Proceedings of the IEEE/CVF conference on computer vision and pattern recognition. pp. 165--174 (2019)

\bibitem{peng2020convolutional}
Peng, S., Niemeyer, M., Mescheder, L., Pollefeys, M., Geiger, A.: Convolutional occupancy networks. In: Computer Vision--ECCV 2020: 16th European Conference, Glasgow, UK, August 23--28, 2020, Proceedings, Part III 16. pp. 523--540. Springer (2020)

\bibitem{previtali2023quantitative}
Previtali, C., Sartoris, R., Rebours, V., Couvelard, A., Cros, J., Sauvanet, A., Cauchy, F., Paradis, V., Vilgrain, V., Burgio, M.D., et~al.: Quantitative imaging predicts pancreatic fatty infiltration on routine ct examination. Diagnostic and Interventional Imaging  (2023)

\bibitem{qi2017pointnet}
Qi, C.R., Su, H., Mo, K., Guibas, L.J.: Pointnet: Deep learning on point sets for 3d classification and segmentation. In: Proceedings of the IEEE conference on computer vision and pattern recognition. pp. 652--660 (2017)

\bibitem{ronneberger2015u}
Ronneberger, O., Fischer, P., Brox, T.: U-net: Convolutional networks for biomedical image segmentation. In: Medical Image Computing and Computer-Assisted Intervention--MICCAI 2015: 18th International Conference, Munich, Germany, October 5-9, 2015, Proceedings, Part III 18. pp. 234--241. Springer (2015)

\bibitem{saito2019pifu}
Saito, S., Huang, Z., Natsume, R., Morishima, S., Kanazawa, A., Li, H.: Pifu: Pixel-aligned implicit function for high-resolution clothed human digitization. In: Proceedings of the IEEE/CVF international conference on computer vision. pp. 2304--2314 (2019)

\bibitem{sun2022topology}
Sun, S., Han, K., Kong, D., Tang, H., Yan, X., Xie, X.: Topology-preserving shape reconstruction and registration via neural diffeomorphic flow. In: Proceedings of the IEEE/CVF Conference on Computer Vision and Pattern Recognition. pp. 20845--20855 (2022)

\bibitem{tewari2020state}
Tewari, A., Fried, O., Thies, J., Sitzmann, V., Lombardi, S., Sunkavalli, K., Martin-Brualla, R., Simon, T., Saragih, J., Nie{\ss}ner, M., et~al.: State of the art on neural rendering. In: Computer Graphics Forum. vol.~39, pp. 701--727. Wiley Online Library (2020)

\bibitem{wasserthal2022totalsegmentator}
Wasserthal, J., Meyer, M., Breit, H.C., Cyriac, J., Yang, S., Segeroth, M.: Totalsegmentator: robust segmentation of 104 anatomical structures in ct images. arXiv preprint arXiv:2208.05868  (2022)

\bibitem{wickramasinghe2020voxel2mesh}
Wickramasinghe, U., Remelli, E., Knott, G., Fua, P.: Voxel2mesh: 3d mesh model generation from volumetric data. In: Medical Image Computing and Computer Assisted Intervention--MICCAI 2020: 23rd International Conference, Lima, Peru, October 4--8, 2020, Proceedings, Part IV 23. pp. 299--308. Springer (2020)

\bibitem{wu20153d}
Wu, Z., Song, S., Khosla, A., Yu, F., Zhang, L., Tang, X., Xiao, J.: 3d shapenets: A deep representation for volumetric shapes. In: Proceedings of the IEEE conference on computer vision and pattern recognition. pp. 1912--1920 (2015)

\bibitem{xie2022neural}
Xie, Y., Takikawa, T., Saito, S., Litany, O., Yan, S., Khan, N., Tombari, F., Tompkin, J., Sitzmann, V., Sridhar, S.: Neural fields in visual computing and beyond. In: Computer Graphics Forum. vol.~41, pp. 641--676. Wiley Online Library (2022)

\bibitem{yang2022neural}
Yang, J., Shi, R., Wickramasinghe, U., Zhu, Q., Ni, B., Fua, P.: Neural annotation refinement: Development of a new 3d dataset for adrenal gland analysis. In: International Conference on Medical Image Computing and Computer-Assisted Intervention. pp. 503--513. Springer (2022)

\end{thebibliography}




% LNCS Homepage, \url{http://www.springer.com/lncs}, last accessed 2023/10/25
% \end{thebibliography}
\end{document}